\documentclass[sigconf]{aamas} 

\usepackage{multirow}
\usepackage{balance} 

\setcopyright{ifaamas}
\acmConference[AAMAS '22]{Proc.\@ of the 21st International Conference
on Autonomous Agents and Multiagent Systems (AAMAS 2022)}{May 9--13, 2022}
{Online}{P.~Faliszewski, V.~Mascardi, C.~Pelachaud,
M.E.~Taylor (eds.)}
\copyrightyear{2022}
\acmYear{2022}
\acmDOI{}
\acmPrice{}
\acmISBN{}

\acmSubmissionID{631}

\title[TAPEM]{Tactile Pose Estimation and Policy Learning for Unknown Object Manipulation}

\author{Tarik Kelestemur}
\affiliation{
  \institution{Northeastern University}
  \city{Boston}
  \state{MA}
  \country{United States}}
\email{kelestemur.t@northeastern.edu}

\author{Robert Platt}
\affiliation{
  \institution{Northeastern University}
  \city{Boston}
  \state{MA}
  \country{United States}}
\email{rplatt@ccs.neu.edu}

\author{Taskin Padir}
\affiliation{
  \institution{Northeastern University}
  \city{Boston}
  \state{MA}
  \country{United States}}
\email{tpadir@northeastern.edu}

\begin{abstract}
Object pose estimation methods allow finding locations of objects in unstructured environments. This is a highly desired skill for autonomous robot manipulation as robots need to estimate the precise poses of the objects in order to manipulate them. In this paper, we investigate the problems of tactile pose estimation and manipulation for category-level objects. Our proposed method uses a Bayes filter with a learned tactile observation model and a deterministic motion model. Later, we train policies using deep reinforcement learning where the agents use the belief estimation from the Bayes filter. Our models are trained in simulation and transferred to the real world. We analyze the reliability and the performance of our framework through a series of simulated and real-world experiments and compare our method to the baseline work. Our results show that the learned tactile observation model can localize the pose of novel objects at 2-mm and 1-degree resolution for position and orientation, respectively. Furthermore, we experiment on a bottle opening task where the gripper needs to reach the desired grasp state.
\end{abstract}

\keywords{tactile pose estimation; tactile manipulation; deep reinforcement learning}

\newcommand{\BibTeX}{\rm B\kern-.05em{\sc i\kern-.025em b}\kern-.08em\TeX}

\begin{document}


\pagestyle{fancy}
\fancyhead{}


\maketitle 


\section{Introduction}
Understanding the physical properties of objects such as their shapes and poses is an essential skill for robotic manipulation in unstructured environments. This problem requires fusing multiple sensor modalities (e.g. vision and touch) in an efficient way. As humans, we use vision and tactile feedback as complimentary sensor modalities to manipulate the objects around us. For example, when we want to use a computer mouse, we first find its rough location on a table using vision and grab it. Then, we locate the buttons on the mouse by using the haptic feedback coming from our fingertips. 

While vision provides rich and global information about the environment, it is difficult to extract precise and local features of the robot-environment interaction. This is particularly important for applications where the robot is constantly in contact with the environment. Moreover, visual sensors are subject to the occlusions by the objects and the robot itself or they might have poor calibration which results in low reliability and performance. In these cases, direct sensory feedback between the robot and the environment is needed. Tactile sensors can determine the contact information regarding the interaction such as whether the robot is touching an object or how much force is applied. 
\begin{figure}[t]
\centering
\includegraphics[width=\linewidth]{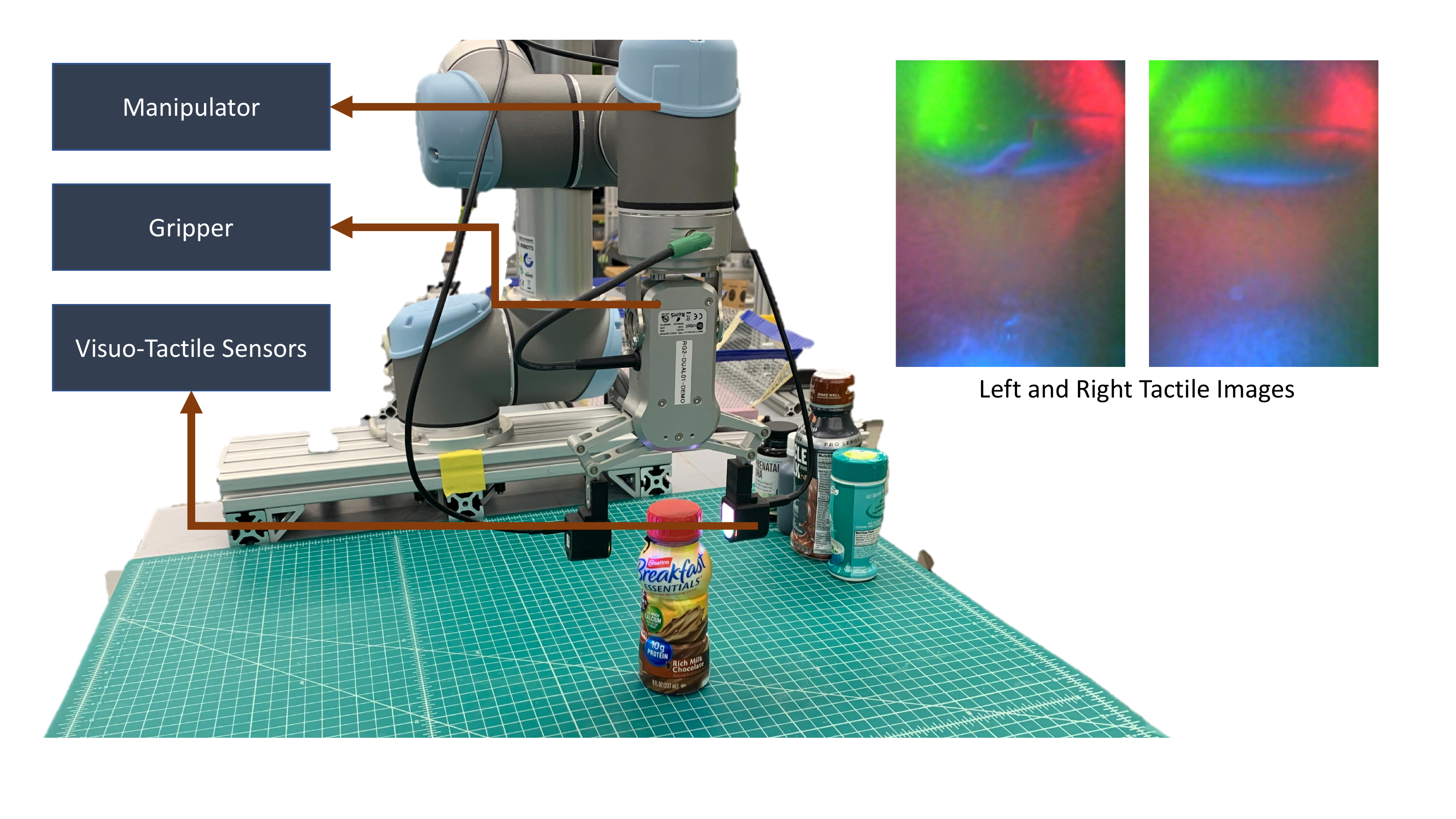}
\caption{Experimental Setup - Our method learns to estimate a robotic gripper pose with respect to manipulated objects. Later, we use this estimator to learn manipulation policies. The experimental setup consists of a UR5 arm, a parallel gripper, and two visuo-tacile sensors mounted on the gripper fingers. The robot grasps the object to perform the pose estimation, then, moves gripper to the desired pose to solve manipulation tasks.}
\end{figure}
In robotic manipulation, before the robot can move an object, it needs to estimate and track the pose of the object in the scene. Once this is achieved, the robot can perform several manipulation tasks. With the rise of deep learning, a line of research~\cite{xiang2017posecnn, wen2021bundletrack} has introduced successful pose estimation methods with vision. These methods train on large datasets of objects with annotated labels and transfer their models to the real world to be used in tasks such as grasping or pushing. However, these methods only provide a rough pose estimation of the objects. Even state-of-the-art methods like~\cite{wen2021bundletrack} can localize the objects at centimeter resolution. Some manipulation tasks (e.g. peg-in-hole, cable insertion, etc.) requires sub-centimeter resolution to be able to achieve high performance. The tactile perception has been shown that it can address this issue, but the majority of the previous work~\cite{li2014localization, luo2015localizing, izatt2017tracking, sodhi2020learning} requires object models to be known or they only work with a limited number of trained objects. Our desire is to go beyond this limitation and be able to precisely localize unseen objects without using any object model.

In this work, we propose a framework for tactile pose estimation and manipulation without any prior knowledge about the objects. Our method first learns an observation model that predicts the state likelihood probabilities of given tactile images. This observation model is then integrated into a discrete Bayes filter with a deterministic motion model. Later, we learn tactile manipulation policies using the belief estimation from this Bayes filter. The tactile observation model is modeled using deep neural networks and learned from data generated in simulation. To learn this model, we build a simulated environment with a parallel gripper and two visuo-tactile sensors as shown in Figure~\ref{fig:sim}. The simulation allows us to collect millions of tactile images with the ground truth data in less than 24 hours. After learning the tactile observation model, we then consider two control problems. The first is the active tactile pose estimation where the robot grasps the object from different states to localize itself in a minimum number of steps. The second problem is to reach a desired goal state on the object. We apply the standard deep reinforcement learning method for solving these tasks. In this setting, the agent takes the belief and goal as input. 

We conduct a series of experiments to show the robustness and performance of our method. In simulation experiments, we first show that the learned tactile observation model can successfully estimate the position of the gripper with respect to the novel objects at 2mm position and 1-degree orientation resolution. Later, we train agents on a bottle opening task where the gripper needs to reach a certain state to open the bottle cap. We compare this method to a common approach where the agent uses recurrent networks instead of belief estimation. Finally, we transfer the tactile observation model and policy networks to a real robot. Our models can work in the real world without extensive domain randomization. We only take a single image from the real sensor to augment the tactile images in simulation. Inspired by the computer mouse example, we envision that this framework can be used as a complimentary estimation method to a vision-based estimator. In this scenario, the vision would estimate the rough pose of the object and our method can precisely localize the with respect to the object. The code and supplementary materials can be found at \href{https://sites.google.com/view/tpem}{https://sites.google.com/view/tpem}.
\section{Related Work}
\subsection{Tactile Pose Estimation}
Pose estimation of objects is an essential skill for autonomous manipulators. While vision-based methods have shown great success in this area, tactile pose estimation is still needed to overcome the shortcomings of the methods that only use visual feedback. For example, visual feedback can be occluded due to arm motions or the hand-eye calibration cannot be always reliable. Moreover, during precise manipulation tasks (e.g. peg-in-hole with low clearance), tactile feedback is shown to be necessary. The previous work by~\cite{bicchi1993contact, gadeyne2001markov, chhatpar2005particle, petrovskaya2006bayesian, corcoran2010measurement, platt2011using, bimbo2015global, saund2017touch} have shown that tactile feedback can be used to estimate objects' poses. The general approach in these works is to formalize the analytical observation and motion models of the tactile interactions and use Bayes filters for pose estimation. A common assumption among these method is that the object is considered to be static. Suresh et al~\cite{suresh2020tactile} considers the object motions and estimates both the object shape and its pose by alternating between Gaussian process implicit surface regression and factor graph. Similarly, Liang et al.~\cite{liang2020hand} takes account of the object motions while it is being held in a multi-fingered hand and localizes the object pose using derivative-free sampling-based optimization. These methods typically use a force sensors that is mounted on the gripper fingers. 

With the recent development of visuo-tactile sensors such as GelSight~\cite{yuan2017gelsight} and DIGIT~\cite{lambeta2020digit}, several work proposed to estimate the pose of grasped objects using visual data. These sensors are equipped with a deformable gel on the contact surface. An RGB camera is placed underneath this gel which captures the deformation cased by contact interactions. Li et al.~\cite{li2014localization} uses a GelSight sensor and apply the RANSAC method to localize a USB connector to perform USB insertion task. A similar approach is taken by Luo et al.~\cite{luo2015localizing} where they estimate the end-effector position by using correspondence matching of SIFT features between the current sensor reading and global features of the surface. Izatt et al.~\cite{izatt2017tracking} uses the data from a GelSight sensor and performs ICP-based tracking with the point cloud data acquired from an external RGB-D sensor. Bauza et al.~\cite{bauza2019tactile} also uses ICP for pose estimation, however, they simultaneously reconstruct the shape of object instead of relying on RGB-D camera. In~\cite{sodhi2020learning}, Sudho et al. learns a tactile observation model that takes visual-based tactile images and uses a factor graph to estimate the relative pose of the sensor while the end-effector is performing planar pushing task. Bauza et al~\cite{bauza2020tactile} proposes an approach to localize objects using a learned observation model that predicts the local shape of the object from visuo-tactile readings. Later, they match this shape to simulated local shapes with known poses using FilterReg algorithm. 

All the methods mentioned above uses some sort of prior knowledge of the objects or only work with a limited number of objects. It is desired to get rid of the this assumption and develop methods that can do pose estimation without the object models. Recently, Anzai and Takahashi~\cite{anzai2020deep} introduced a method called deep gated multi-modal learning for fusing visual and tactile observations to estimate the pose changes of different objects. Their method takes tactile data from a GelSight sensor and visual data from an RGB camera placed on the end-effector. The visual and tactile data are encoded using convolutional and recurrent layers. The networks are trained on 11 objects and tested on 4 novel objects. Contrast to their work, we do not use any camera images and solely rely on tactile images. Moreover, our method estimate the actual pose of the object instead of the relative changes. Finally, we propose to use this estimation method for solving manipulating tasks using deep reinforcement learning.

\subsection{Tactile Manipulation}
The main goal of tactile manipulation is to close the loop for solving manipulation tasks. It has been shown by several works that tactile information is not only useful but also necessary to solve precise manipulation problems. In~\cite{lee2020making}, Lee et al. investigate the representation learning for visual and tactile fusion for manipulation. They use a Variational Autoencoder to learn a compact representation of these two modalities which is then used for learning the peg-in-hole task with unseen shapes and sizes. She, Wang, and Dong et al.~\cite{she2019cable} uses a GelSight sensor to estimate the pose of a gripped cable and friction forces during a cable sliding task. Wang and Wang et al.~\cite{9341006} learns a physical feature embedding space by performing exploratory actions. This embedding is then used to predict the swing angle for the given control input. Dong et al~\cite{dong2021tactile} proposed to use vision-based tactile images to learn end-to-end policies for insertion tasks in the real world. A sim-to-real approach for tactile manipulation is produced by~\cite{ding2021sim}. They calibrate the tactile sensor in the simulation with real-world data and learn door opening tasks using deep reinforcement learning. 

While we show that end-to-end policy learning is possible for the tactile manipulation tasks considered in this work, decoupling the perception and control problems results in faster learning and better generalization to unseen objects. It is also important to note that, with a learned tactile pose estimator, various tasks can be easily learned instead of learning each task from scratch.

\subsection{Partial Observability in Deep Reinforcement Learning}
In many real-world scenarios, the robots do not have access to the underlying state of their environment, instead, they perceive the world through the sensors that provide partial observations. While deep reinforcement learning (DRL) has been successfully applied to robotics tasks with full state information, policy learning under partial observability is still an ongoing research effort. To address this problem, the previous work has proposed two main solutions. The first is to deploy a memory module such as Long Short Term Memory (LSTM)~\cite{hochreiter1997long} to the value and policy networks. The work by Hausknecht et al.~\cite{hausknecht2015deep} proposed the Deep Recurrent Q-Network which modifies the original Deep Q-Network~\cite{mnih2015human} architecture by adding LSTM layers to the network. Later, this approach is also applied to the actor-critic methods~\cite{espeholt2018impala} and model-based methods~\cite{ha2018world}. 

The second approach is to take a modular approach where the state estimation is realized by differentiable Bayes filters and the policy takes the belief estimates as the input. In partially observable environments, a Bayes filter~\cite{cifuentes2016probabilistic} maintains a posterior probability (belief) over the states and recursively updates its belief using the observation and transition models. The recent progress in neural networks has shown that these models can be learned from data if their analytical forms do not exist. A differentiable discrete Bayes filter is proposed by Jonschkowski et al.~\cite{Jonschkowski-16-NIPS-WS} and showed that it outperforms LSTM networks in state estimation problems. The work by Karkus et al.~\cite{karkus2018particle} and Jonschkowski et al.~\cite{Jonschkowski-RSS-18} learns Particle filters and the work by Haarnoja et al.~\cite{haarnoja2016backprop} learns a Kalman filter to handle continuous states. Lee and Yi et al.~\cite{lee2020making} showed how to fuse different sensor modalities while learning Bayes filters. 

Combined with the DRL, the differentiable Bayes filters give promising results in partially observable robotic environments. In~\cite{karkus2017qmdp}, Karkus et al. introduce QMDP-net which combines the QMDP planner with a Histogram filter and jointly learns them to solve simulated navigation and grasping tasks. Wirnshofer et al.~\cite{Wirnshofer-RSS-20} uses a learned particle filter to track objects in the environment and trained a DQN agent to manipulate these objects. The agent uses the belief estimates as input to the DQN. Chaplot et al~\cite{singh2018active} and Gottipati et al.~\cite{gottipati2019deep} applied this approach to the active localization problem with visual feedback. Finally, in~\cite{kelestemur2020learning}, the authors proposed to use a learnable Bayes filter to localize a robotic gripper's position with respect to the environment image using tactile feedback. In the following work~\cite{kelestemur2021policy}, they train agents using deep reinforcement learning with belief inputs to solve contact-rich manipulation tasks using only a single image of the environment and the tactile observations.

In this work, we take the latter approach where we train an observation model which provides likelihood probabilities of the tactile observations which are later used within a discrete Bayes filter. Then, we train DRL agents for estimating the gripper's pose with respect to unknown objects and manipulate them. Our experiments and results presented in related work suggest that learning policies with Bayes filters result in better performance compared to the policies learned with recurrent networks.
\section{Background}
\noindent
\subsection{Partial Observability and Bayes Filters}
We formulate our pose estimation problem as a partially observable Markov Decision Process (POMDP)~\cite{kaelbling1998planning} which is defined with a tuple $(\mathcal{S}, \mathcal{A},\Omega, \mathcal{T}, \mathcal{R}, \mathcal{O})$ where $\mathcal{S}, \mathcal{A}$, and $\Omega$ are the state, action, and observation spaces, respectively. $\mathcal{T}=p(s_t|s_{t-1}, a_{t-1})$ is the state-action transition function that gives next state probabilities given the action $a_{t-1} \in\mathcal{A}$ and previous state $s_{t-1}\in\mathcal{S}$. The observation model $\mathcal{O} = p(o_t|s_t)$ defines the likelihood probabilities of receiving the observation $o_t\in\Omega$ at state $s_t$. At each timestep, the agent also receives a reward $r_t\in\mathbb{R}$ provided by the reward function $\mathcal{R}(s_t)$. 

A Bayes filter is a recursive state estimation method that can be used when agents do not have full access to systems' states. To estimate the current state of the agent, Bayes filters maintain a belief which describes the posterior probability over state space. This belief is conditioned on the past observations and the actions $bel(s_t) = p(s_t| a_{1:t-1}, o_{1:t})$ and recursively updated at every time step using the observation and transition models:
\begin{align}
    \overline{bel}(s_t) &= \underbrace{\sum_{s_{t-1} \in  S}\mathcal{T}(s_t, a_{t-1}, s_{t-1})bel(s_{t-1})}_{\textit{Prediction Update}}\label{eq:prediction_update}\\
    bel(s_t) &= \underbrace{\eta\mathcal{O}(o_t, s_t)\overline{bel}(s_t)}_{\textit{Observation Update}}\label{eq:observation_update}
\end{align}
where $\overline{bel}(s_t)$ is the predicted belief and $\eta$ is the normalization factor. The Eq.~\ref{eq:prediction_update} predicts the belief at next time step $\overline{bel}(s_t)$ by summing over the all possible previous states. Then, in Eq.~\ref{eq:observation_update}, the belief is updated using the observation received at the new state. The recursive estimation starts with a prior belief $bel(s_0)$ which is initialized uniformly if there is no knowledge about the system's state.

\subsection{Reinforcement Learning}
In partially observable environments where we have belief estimation, the environment becomes a belief Markov Decision Process (BMDP).  \textbf{Belief MDP} is defined as a tuple $(\mathcal{B}, \mathcal{A}, \mathcal{R}, \tau, \gamma)$ where $\mathcal{B}$ is the belief space, $\mathcal{A}$ is the action space and $\mathcal{R}:\mathcal{B}\times \mathcal{A}\rightarrow\mathbb{R}$ is the reward function. The $\gamma$ is the discount factor and $\tau:\mathcal{B}\times\mathcal{A}\rightarrow\mathcal{B}$ is belief-transition function. In this formulation, the action space $\mathcal{A}$ is same with the underlying POMDP. In discrete state spaces, the belief is defined as a categorical distribution. 
\begin{figure}[t]
\centering
\includegraphics[width=0.98\linewidth]{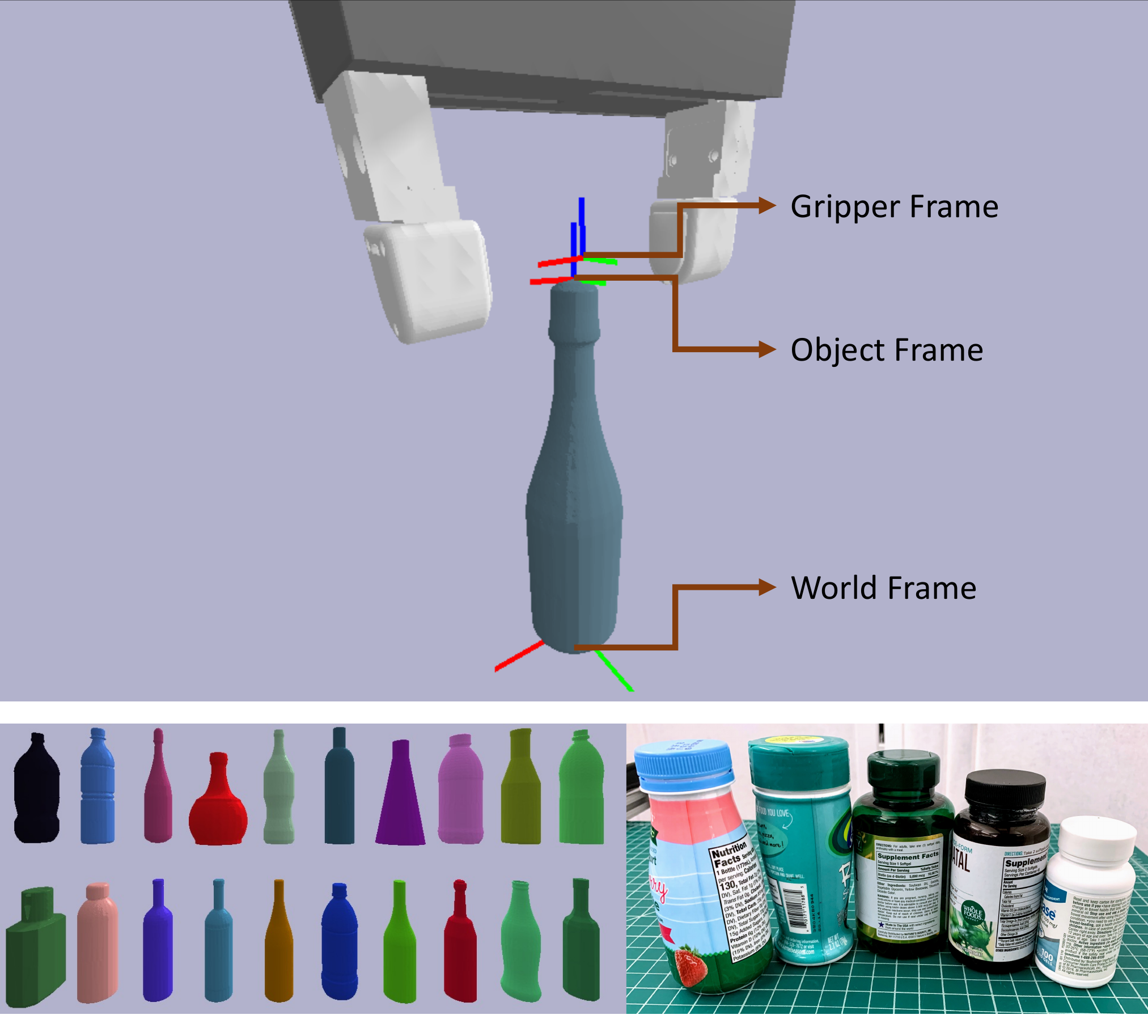}
\caption{\textit{Top Image}: Simulation environment - \textit{Bottom-left Image}: Examples from simulated objects - \textit{Bottom-right Image}: Real-world objects}
\label{fig:sim}
\end{figure}

\textbf{Goal conditioned RL} (GCRL)~\cite{kaelbling1993learning} allows agents to learn policies that can reach desired goals. The goal of GCRL is to find a policy $\pi(a_t|s_t, g)$ that is conditioned on the state and the goal and maximize the expected discounted return: $\mathbb{E}_{\pi} [\sum_{t=0}^{T-1}\gamma^{t}\mathcal{R}(s_t, g)]$ where $T$ is the maximum horizon and, $\gamma\in[0,1]$ is the discount factor, and $\mathcal{R}(s_t, g)$ is the goal-conditioned reward function. At the beginning of each episode, an initial state $s_0\sim p_0$ and a goal $g\sim p_g$ are sampled from the state and goal distributions. In the case of BMDP, the policies and the reward function are conditioned on the belief instead of the underlying state: $\pi(a_t|bel(s_t), g)$ and $R(bel(s_t), g)$. With this change, the goals also need to be in the belief space $g\in\mathcal{B}$. The goals can be represented as one-hot vectors which in the belief space corresponds to a probability distribution where the probability of goal state is 1 and the rest is 0.

\section{Methods}
\subsection{Problem Statement} \label{sec:prob_statement}
In this paper, we tackle two problems: estimating a gripper's pose with respect to an object using tactile feedback and moving the gripper to the desired pose on the object to achieve a manipulation task. Below, we explain how these problems are formulated.

\textbf{Tactile Pose Estimation:}
To estimate the pose of a gripper with respect to the objects, we use a discrete Bayes filter (or Histogram filter) with a learned observation model and a deterministic transition function. We represent the observation model as a neural network and learn it from data generated in simulation. Formally, the observation model is a function that takes the tactile images from two fingers and produces state likelihood probabilities: $\mathcal{O}(s_t, o_t)=h_\theta(I_l, I_r)$ where $\theta$ is the network parameters, $I_l$ and $I_r$ are the tactile images acquired from left and right fingers. The tactile observations are 3-channel $h\times w$ RGB images $I_l, Ir\in\Omega=\mathbb{Z}^{3\times h\times w}$. The belief-transition function takes the previous state and the action to output the next state $\tau:f(bel(s_{t-1}), a_{t-1})=\overline{bel}(s_t)$. 

Our state space is described as the position and orientation of the gripper with respect to the object. To simplify things, we place the objects into the center of the world frame and assign the object's frame to the highest point on the object in the z (upward) axis while keeping the orientation the same as the world frame. The gripper's frame is positioned in the middle of the tactile sensors (see Figure~\ref{fig:sim}). We discretize the positions and orientations and represent them as state bins where the middle element of bins represent the origin pose. The discretization resolution for the position is 2mm and for orientation, it is 1 degree. Since the discretization in each position and orientation can easily create a large state space, we use a factored state representation where the state $s$ is a $n\times d$ matrix $s\in \mathcal{S}=\mathbb{Z}^{n\times d}$ where $n$ is the state dimension and $d$ is the size of that dimension.

Similar to the state space definition, the belief is also represented as $n\times d$ matrix $bel(s)\in\mathcal{B}=\mathbb{R}^{n\times d}$. Because we use a factored belief representation, each row of the belief is a probability mass function $bel(s^i)~\forall i=1,..,n$ for the corresponding discretized position or orientation space. To do recursive belief updates, we first shift the belief by the distance the gripper moved and then apply element-wise multiplication with the output of the observation model:
\begin{align}
 bel(s_t)=\eta h_\theta(I_l, Ir)\odot f(bel(s_{t-1}), a_{t-1})   
\end{align}
The initial belief $bel(s_0)$ is uniform in each state dimension.

\textbf{Tactile Manipulation:}
Once we are able to localize the gripper's pose, our next goal is to move the gripper to the desired pose to solve a manipulation task. To this end, we learn a goal-conditioned policy that takes the the current belief $bel(s_t)$ and the goal $g$ to move the gripper $\pi(a_t|bel(s_t), g)$. Moreover, we learn a policy that can estimate the gripper's pose in minimum number steps $\pi(a_t|bel(s_t))$ rather than taking random actions. It is important to state that the localization and manipulation policies are trained separately and the manipulation policy does not require the localization to be learned. Instead, it implicitly learns to localize gripper to reach the goal.
\begin{figure*}[t]
    \centering
    \includegraphics[trim={0cm 1cm 0cm 0cm}, width=0.99\linewidth]{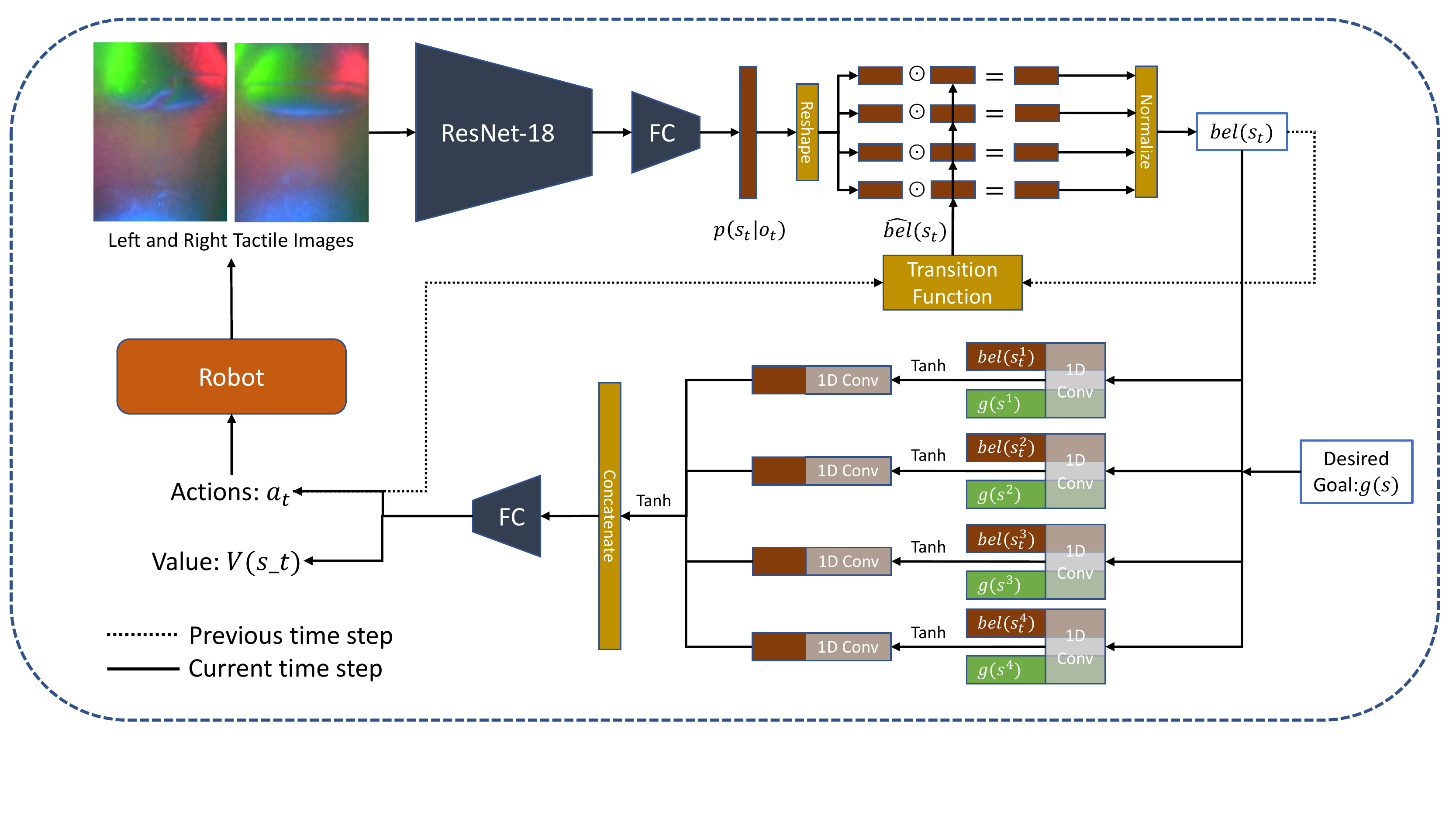}
    \caption{The System Framework -- Our proposed system has two main components: (a) a factored Bayes filter with a learned observation model and (b) a policy-value network that uses belief estimations to reach desired goals. The observation model uses a ResNet network to encode tactile images and produces the state-likelihood probabilities. The deterministic transition model predicts the belief at the next time step which is then multiplied and normalized with the likelihood probabilities. Finally, the policy network takes the belief along with the goal to output actions.}
    \label{fig:network}
\end{figure*}
\subsection{Tactile Observation Model} \label{sec:tactile_obs_model}
\textbf{Architecture and Training:} For the observation model, which we call Tactile Pose Network (TPN), we use the ResNet-18~\cite{he2016deep} architecture pre-trained on the ImageNet~\cite{deng2009imagenet} dataset. The final fully connected (FC) layer is removed and replaced with two FC layers which have 2048 and $n\times d$ hidden units. The first layer is followed by a ReLU activation and the second one is normalized using Softmax. Since we have two images for each state, we call the forward pass twice and accumulate the gradients during training. The tactile images are resized to $85\times 64$ from $160\times120$. To train the network, the categorical cross-entropy loss is used. Since we are using a factored state representation, the loss is calculated as the sum of negative log-likelihoods over the state dimensions: 
\begin{align}
    \mathcal{L} = -\sum_{i}^{n}\sum_{j}^{d}s_j^i\log(h_\theta(I_r, I_l))
\end{align}
where $s_j^i$ is the ground-truth state at $i-th$ dimension and $j-th$ position in the state dimension. We use Adam optimizer with a learning rate of 0.0005. The batch size is 64 and the number of epochs is 20. The training takes 2 hours. The overall architecture can be seen in Figure~\ref{fig:network}. 

\textbf{Simulation and Data Collection:} We build a simulation environment using the PyBullet~\cite{coumans2021} library as the physics engine and the TACTO~\cite{wang2020tacto} as the renderer for the visuotactile sensors. TACTO is a recent framework that allows simulating visuotactile sensors. It currently supports two sensors: DIGIT~\cite{lambeta2020digit} and OmniTact~\cite{padmanabha2020omnitact}. In our simulation and real-world experiments, we use DIGIT sensor due to its open-source hardware and software implementation. We placed two DIGIT sensors on the fingertips of a parallel gripper. The simulation is loaded with 3D objects that are acquired from ShapeNet~\cite{shapenet2015} dataset and pre-processed using~\cite{huang2020manifoldplus} to make them watertight. We found that this process improves the quality of tactile images. To collect the dataset, the gripper is placed in every state and the fingers are closed with a position controller with a limited maximum force. The dataset is populated with the pair of tactile images (left and right fingers) and the corresponding state. The data collection takes approximately 24 hours. 

\textbf{Bayes Filter:} With the observation model is learned, we then integrate it into a discrete Bayes filter. The output of the observation model gives the likelihood probabilities for the given tactile images: $p(s_t|I_r, I_l)$. To do the factored belief updates, we first reshape the output of TPN from a vector with size $nd$ to a matrix with shape $n\times d$ where each row represented likelihoods for the corresponding state dimension. Then, each row is multiplied with the predicted belief $\overline{bel}(s_t)$ element-wise to generate the belief at the next time step. The predicted belief is produced by the deterministic belief-transition function $f(bel(s_{t-1}, a_{t-1})$ which takes the previous belief and the action and shifts the belief by the distance the gripper moved.
\subsection{Policy Learning}
\textbf{Tasks:} To train and test the policies, we developed a simulated task environment using Gym~\cite{gym} interface. At the beginning of each episode, the gripper is placed in a random pose and the belief is initialized uniformly. At each timestep, the gripper moves to a new pose according to the action requested and grasps the object to collect the tactile images. These images are then returned back to the agent. We use the same bottle dataset collected for the observation model. Here, we are interested in learning two tasks: \textit{Active Pose Estimation} where the agent takes actions to localize itself and \textit{Reaching} where the gripper is trying to reach the desired goal pose.

\textbf{Rewards:} For both of the tasks, we use sparse reward functions. The reward function for active pose estimation task is $\mathcal{R}(s_t)=1$ \textit{if} $|s_t-\hat{s_t}|=0$ \textit{else} $0$ where $s_t$ is the current true state and $\hat{s_t}$ is the estimated state. Here, the estimated state is defined as the point in the belief with highest probability value $\hat{s_t}=\textit{argmax}(bel(s_t))$. For the reaching task, the reward function is $\mathcal{R}(s_t, g)=1$ \textit{if} $|s_t-g|=0$ \textit{else} 0 where $g$ is the goal state. We terminate the episode if the reward is 1 or the maximum time limit is reached. For both tasks, the task horizon is 16.

\textbf{Policy Architecture:} Proximal Policy Optimization (PPO)~\cite{schulman2017proximal} is an actor-critic RL method that directly learns a policy network as well as a value network. Different from other actor-critic methods, PPO limits the updates to the policy network by clipping the policy gradient objective. This leads to a low-variance and stable learning. The value and policy networks are consists of 2 1-d convolutional layers and a single FC layer. Both of the convolutional layers have one kernel with size 3 and they are followed by the Tanh activation function. The FC layer for the policy network has 20 units whereas the value FC layer has 1 unit. The weights of the convolutional layers are shared between the policy and the value networks. The agent takes the belief as input to the networks. In the case of learning a manipulation task, it also takes the goal which is represented as one-hot vectors. For each state dimension, we stack the belief and the goal as two-channel vectors and feed them into the network. We found out that using convolutional layers works better than fully connected layers because the input to the network has spatial properties. The architecture for the policy and value network can be seen in Figure~\ref{fig:network}. 

\textbf{Multi Dimensional Discrete Action Space:}
Our policies use a discrete action space. Similar to the belief space, the action space is has a factored representation. In the case of discrete action space, the policy $\pi(\cdot|bel(s))$ is represented as a softmax distribution. If we have a multi-dimensional action space, we can specify categorical distribution $\pi_{\theta_i}(a_i|bel(s))$ over actions $a_i\in \mathcal{A}_i$ for each dimension $i$. Then, we can define the joint discrete policy as $\pi(a|bel(s)):=\Pi_{i=1}^{m} \pi_{\theta_{i}}\left(a_{i}| bel(s)\right)$ where $m$ is the dimension of the action space. To implement this, the policy network has a fully connected layer with the size of $\sum_{i=1}^mK_i$ where $K_i=|\mathcal{A}_i|$. Then, we would apply softmax to each action space dimension. We use delta position commands for controlling the gripper. For each action dimension, the agent can move $\pm1$, $\pm2$ elements in the discrete state space or stay still. We have 4 state dimensions, thus, in total, we have 20 discrete actions. In position dimensions, each bin corresponds to a 2mm change and for the orientation dimension, it is a 1-degree change.
\begin{table}[t]
    \caption{TOP-5 OBSERVATION MODEL ACCURACY}
    \begin{center}
        \begin{tabular}{ c  c | c | c | c | c }
            \hline
            \multicolumn{6}{c}{Validation Set}\\
            \midrule
            State  & top-1 & top-2 & top-3 & top-4 & top-5\\
            \hline
            Position Y & 99.98 & 100.0 & 100.0 & 100.0 & 100.0\\
            \hline
            Position Z & 99.99 & 100.0 & 100.0 & 100.0 & 100.0\\
            \hline
            Rotation X & 98.85 & 99.96 & 99.99 & 100.0 & 100.0\\
            \hline
            Rotation Y & 98.32 & 99.78 & 99.94 & 99.96 & 99.97\\
            \hline
            \hline
            \multicolumn{6}{c}{Holdout Set}\\
            \hline
            State & top-1 & top-2 & top-3 & top-4 & top-5\\
            \hline
            Position Y & 99.00 & 99.92 & 99.96 & 99.98 & 100.0\\
            \hline
            Position Z & 86.83 & 91.42 & 94.20 & 96.32 & 97.50\\
            \hline
            Rotation X & 75.71 & 93.50 & 97.61 & 99.04 & 99.56\\
            \hline
            Rotation Y & 76.43 & 92.52 & 96.51 & 98.02 & 98.67\\
            \hline
        \end{tabular}
    \end{center}
  \label{tab:topk}
\end{table}
\section{Experiments}
To evaluate our method, we first investigate the performance of the learned Bayes filter and policies in simulation. Later, we transfer our models trained in simulation to the real world. As explained in Section~\ref{sec:prob_statement}, we are interested in estimating the pose of a gripper with the respect to manipulated objects and control the gripper to move to the desired pose. To this end, we choose the bottle opening task where the gripper needs to find its pose with respect to a bottle and move to appropriate to the pose to twist and open the bottle cap. Here, we have 3 expectations from our method: (I) the pose estimation should be robust and generalize to unseen objects, (II) the localization and manipulation policies should learn effective control strategies, (III) our models should be able to transfer to the real world without fine-tuning or intensive domain randomization. 
\begin{figure*}[t]
    \centering
    \includegraphics[width=0.99\linewidth]{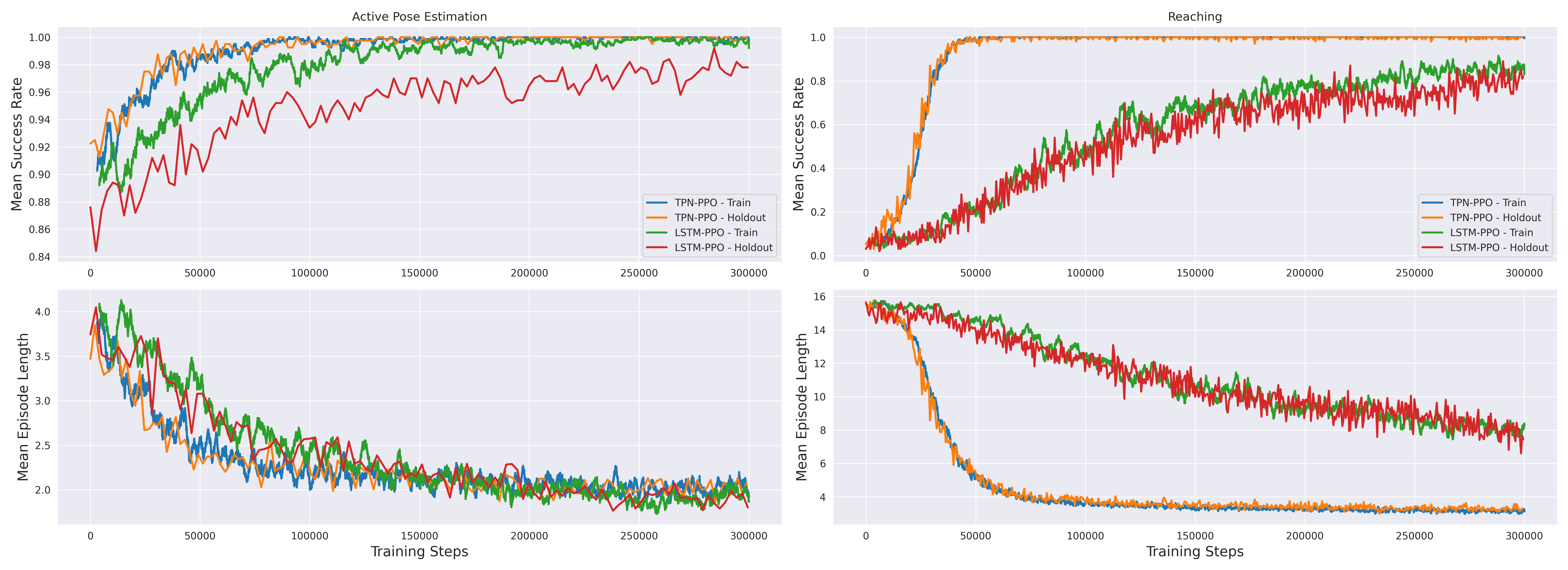}
    \caption{The Policy Learning Curves -- This figure shows the success rates and mean episode lengths over the course of policy learning. We train the policies on a dataset with 50 bottles and evaluate during the training with 10 unseen bottles. The training is averaged over 4 different random seeds. The policies trained with belief estimation outperforms the policies with the recurrent network in terms of sample-efficiency and final performance.}
    \label{fig:ppo}
\end{figure*}
\subsection{Simulation Experiments}
\textbf{Tactile Pose Network:} To train the TPN model, we selected 60 bottles from the ShapeNet dataset. 50 of these bottles are used for the train/validation set and 10 of them are used as the holdout set. We do a 90/10 percent split for the train/validation set. For the bottles dataset, we used 4 dimensions $n=4$ where the first two is the gripper's position in y and z axis and the last two are the orientation around the x and y axes. While we use 4 dimensions in this experiment, the number of the dimensions can be easily increased for tasks that requires more than 4 dimensions. We chose the size of each dimension to be $d=11$. For the position dimensions, this corresponds to 2cm and for orientation it is 10-degrees. 

After training is completed, we run the model on both the validation and holdout sets and report top-5 accuracy in Table~\ref{tab:topk} each state dimension. The results show that the observation model not only predicts the true state (99.28\% mean top-1 validation accuracy) successfully but also generalizes to the novel objects that were not present in the training set (84.49\% mean top-1 holdout accuracy). One can say that this model can be directly used to predict the states and learn policies with state input to the agent instead of the belief estimation. While this is true in the simulation, the belief estimations are going to be essential when noise is introduced into the system which is highly likely in the real-world case. Moreover, using uncertainty as an input to policy learning allows agents to trade between taking exploratory actions and actions towards the goal.

\textbf{Policy Performance}: We train the active pose estimation and reaching tasks with the training set used in the observation model and evaluate them using the holdout set. During training, the evaluation policy is executed for 100 episodes after each gradient update to the agent networks. We compare our method (called \textit{TPN-PPO}) to a baseline where we do not use the Bayes filter, instead, the agent uses an LSTM layer for handling the partial observability. This baseline (called \textit{LSTM-PPO}) takes the tactile observations directly and feeds them into the ResNet-18 network. Similar to the TPN, this ResNet-18 is also pre-trained on the ImageNet dataset. The last FC layer of the ResNet architecture is followed by a ReLu activation. To encode the goal state, we use a single FC layer with 128 hidden units which are followed by a ReLu. The feature vectors from the ResNet encoder and goal encoder are then concatenated and fed into a single layer LSTM layer with a hidden size of 1024. The LSTM layer is followed by ReLu activation and another FC layer with 512 hidden units. Finally, the output of this layer is fed into two FC layers representing the policy and value networks. The weights of the network except the last FC layers are shared between value and policy networks. We report the mean success rate and mean episode lengths over the course of training in Figure~\ref{fig:ppo}. The learning curves are averaged over 4 runs with different random seeds. The policy learning curves show that TPN-PPO outperforms the LSTM-PPO significantly in terms of both learning time and the final success rate. Moreover, it is clear that the policies are able to solve the tasks with the unseen objects as well as the training objects. The overall training for the TPN-PPO (including the observation model) takes approximately 4 hours whereas the training the LSTM-PPO takes 6 hours.
\subsection{Robot Experiments:}
We transfer the observation model and policy networks to the real world to evaluate our method on a real robot. An OnRobot RG2 parallel gripper is attached to a UR5 arm and two DIGIT sensors are mounted on the finger of the gripper. The reason we can effectively transfer our models to real world is that, we augment the tactile images in simulation with the background image taken from the real sensor. This calibration is realized by the TACTO simulator where it calculates the pixel-wise difference before and after the touch and adding it to the background from the real sensor. We also add Gaussian noise to the tactile images during the training of the TPN to make the real-world predictions more robust. Another reason is that due to the sequential nature of the recursive belief estimation, the belief of pose is corrected in the consecutive timesptes, if observation module made a wrong prediction. 
\begin{table}[t]
    \caption{SUCCESS RATE \& MEAN EPISODE LENGTHS}
    \begin{center}
        \begin{tabular}{ c  c | c | c | c | c }
            \hline
            Bottles & 1 & 2 & 3 & 4 & 5\\
            \midrule
            Success Rate & 16/20 & 17/20 & 16/20 & 14/20 & 18/20\\
            \hline
            Mean Episode Length & 8.57 & 6.42 & 5.47 & 8.19 & 7.09\\
            \hline
        \end{tabular}
    \end{center}
  \label{tab:real_worl_results}
\end{table}
In real-world experiments, the task for the robot is to align the gripper with the bottles' cap and open it. Here, the goal is defined as the origin pose of the bottle cap. Once the gripper reaches this pose, it simply rotates the last link of the manipulator to open the cap. The actions for the robots are the same as the simulation i.e. move the end-effector in each state dimension by a fixed distance. To execute the actions, we solve the inverse kinematic problem and send the joint angles to the position controller running on the UR5. We used 5 bottles which are placed on a table. Note that the models of the real world bottles are not used in the training set. For each object, we run the policy 20 times starting from different initial poses. Even though the optimal policy can solve this task on an average of 4 steps in the simulation, due to noise from the real tactile readings, the policy takes longer to solve in the real world. The robot is allowed to take 20 actions before we terminate the episode. After each action is taken, we look at the error between estimated belief and the goal state $e=|\hat{s}_t-g|$ and terminate the episode if the error is less than 3. The episode is considered successful if the gripper was able to rotate the cap. We want to note that even though we used a fixed goal pose for this task, the reaching policy is able to reach any pose in the state space. This allows solving sequential manipulation tasks where the gripper needs to reach multiple goals.

We report the success rate and the mean episode lengths for each object in Table~\ref{tab:real_worl_results}. As can be seen from the results, the robot is able to solve this task efficiently in a real-world setting with $81\%$ success rate. A failure case we identified is that in some cases the policy moves the gripper in a state where the gripper cannot grasp the bottle cap, instead, fully closes the gripper. This results in wrong prediction for the observation model and the policy can not recover. 
\section{Conclusions}
In this work, we introduce a method that can estimate the pose of a robotic gripper with respect to objects and manipulate them using tactile observations. We achieve this without using prior knowledge about the shape or size of the objects which is desired in real-world settings. The tactile pose estimation is realized via a factored Bayes filter with a learned observation model. Furthermore, we learn policies that can use this belief estimation for solving manipulation tasks. The performance and robustness of our method are analyzed through experiments conducted in simulation and the real world. 

The main limitation of our work is the assumption that the manipulated objects are considered to be static during the robot interactions. While this assumption holds for some objects in the real world, ideally, we would like our method to consider the motion of the object. This would also allow objects to be manipulated without re-grasping. For future work, we also want to explore different manipulation tasks where our methods can be applied. Moreover, we would like to go beyond the category-level objects and develop a tactile observation model that can localize the gripper with respect to multiple categories of objects. Finally, to further improve the real-world results, we are planning to explore sim-to-real methods that could be applied to the visuo-tactile images for learning policies with visual inputs.
\begin{figure}[t]
    \centering
    \includegraphics[trim={2cm 0cm 2cm 0cm}, width=0.98\linewidth]{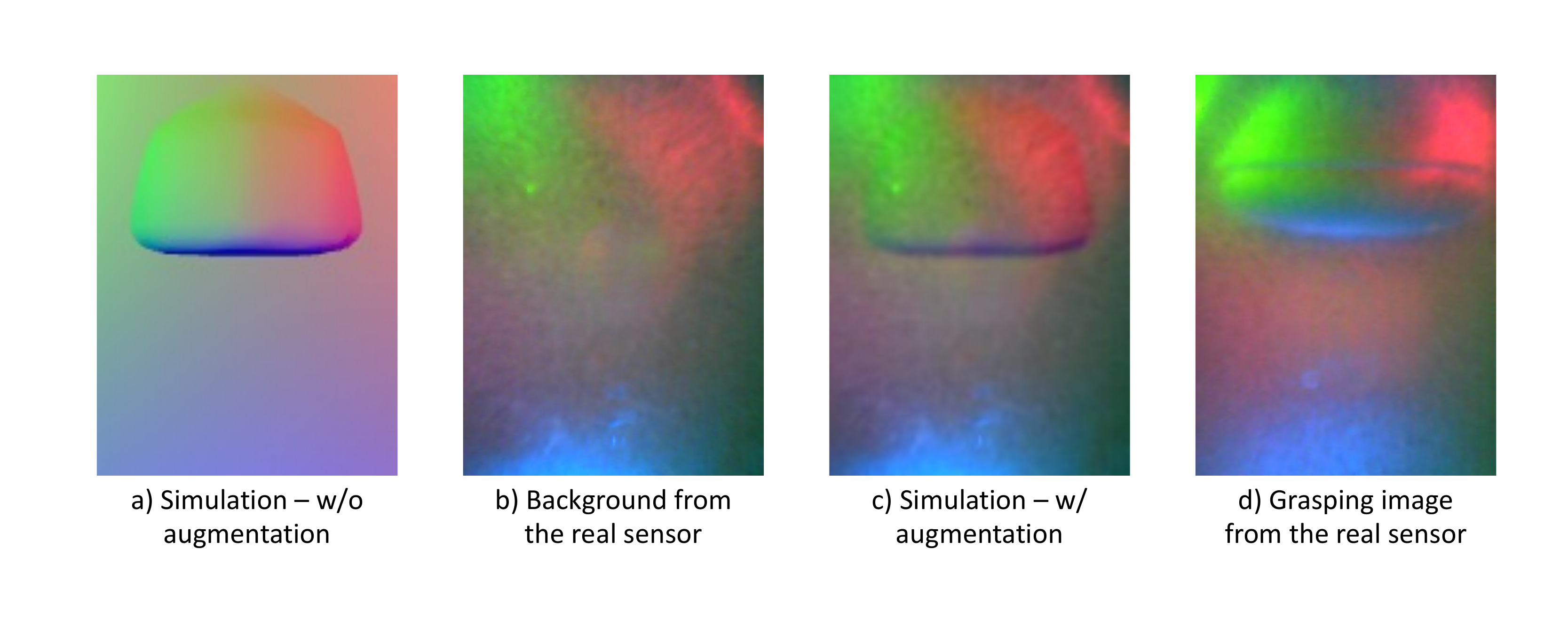}
    \caption{Tactile Image Augmentation - We augment the images in the simulation with the background image of the real sensor for improving the sim-to-real performance.}
    \label{fig:ppo}
\end{figure}
\balance



\begin{acks}
\texttt{This research is supported by the National Science Foundation under Award
Number $1928654$ and the U.S. Office of Naval Research under award number $N00014-19-1-2131$.}

\texttt{The authors would like thank Iris Wang, Joel Willick and Hillel Hochsztein for their help with manufacturing the DIGIT sensor used in this work, and Ondrej Biza for his insights on the neural network design for policy learning.}
\end{acks}



\bibliographystyle{ACM-Reference-Format} 
\bibliography{references}
\end{document}